\pdfoutput=1

\documentclass[11pt]{article}

\usepackage[final]{acl}

\usepackage{times}
\usepackage{latexsym}

\usepackage{arabtex}
\usepackage{utf8}

\usepackage[T1]{fontenc}

\usepackage[utf8]{inputenc}

\usepackage{microtype}

\usepackage{inconsolata}

\usepackage{graphicx}

\usepackage{tipa}

\title{Exploiting Dialect Identification \\
in Automatic Dialectal Text Normalization}

\author{Bashar Alhafni, Sarah Al-Towaity, Ziyad Fawzy, Fatema Nassar, Fadhl Eryani,
\\
\textbf{Houda Bouamor}\textsuperscript{\textdagger}, \textbf{Nizar Habash}\\
  Computational Approaches to Modeling Language Lab\\
  New York University Abu Dhabi\\
  \textsuperscript{\textdagger}Carnegie Mellon University in Qatar\\
  \texttt{\{alhafni,sa5793,fan6236,za2051,fadhl.eryani,nizar.habash\}@nyu.edu}\\
  \texttt{hbouamor@cmu.edu}
  }

\newcommand{\hide}[1]{}

\newcommand{\TAMARBUTA}{{$\hbar$}}

\newcommand{\THA}{{$\theta$}}

\newcommand{\SHIN}{{\v{s}}}

\newcommand{\GAYN}{{$\gamma$}}

\begin{document}
\setcode{utf8}
\maketitle
\begin{abstract}
Dialectal Arabic is the primary spoken language used by native Arabic speakers in daily communication. The rise of social media platforms has notably expanded its use as a written language. However, Arabic dialects do not have standard orthographies. This, combined with the inherent noise in user-generated content on social media, presents a major challenge to NLP applications dealing with Dialectal Arabic. In this paper, we explore and report on the task of CODAfication, which aims to normalize Dialectal Arabic into the Conventional Orthography for Dialectal Arabic (CODA). We work with a unique parallel corpus of multiple Arabic dialects focusing on five major city dialects. We benchmark newly developed pretrained sequence-to-sequence models on the task of CODAfication. We further show that using dialect identification information improves the performance across all dialects. We make our code, data, and
pretrained models publicly available.\footnote{\url{https://github.com/CAMeL-Lab/codafication}}
\end{list}
\end{abstract}

\section{Introduction}

Arabic exhibits a diglossic \cite{Ferguson:1959:diglossia} linguistic situation where a non-standard variety, Dialectal Arabic (DA), coexists with Modern Standard Arabic (MSA), the standard form of the language. Complicating matters, there are multiple DA varieties, each differing from both other dialects and MSA in phonology, morphology, and lexicon. Arabic dialects are typically classified regionally, e.g., Egyptian, North African, Levantine, and Gulf.
These dialects are the true native languages historically connected to Classical Arabic and other regional languages. While Arabic dialects are primarily spoken, they are increasingly used in written form on social media. Since Arabic dialects lack standard orthographies \cite{Habash:2018:unified}, DA text tends to be highly varied and noisy.

This high degree of noise poses major challenges for NLP systems as it increases the degree of sparsity in the data. Such noise can be handled using modeling techniques that normalize DA if it is used as an input to the system, e.g., in machine translation from dialects to other
languages. However, challenges arise when the dialect itself is the desired output, for example, in automatic speech recognition systems \cite{Ali:2019:towards,sahyoun-shehata-2023-aradiawer}. Consequently, evaluating and optimizing these systems can become problematic.

To mitigate the lack of orthographic standards for DA, several efforts in Arabic NLP introduced a common convention for DA spelling, named Conventional Orthography for Dialectal Arabic (CODA) \cite{Habash:2018:unified}.
However, the majority of approaches involving CODA consider it a side task to efforts like morphological disambiguation, diacritization, and lemmatization, as opposed to being the main target task. 

Our contributions in this paper are as follows:
\begin{itemize}
    \item We explore and report on the task of CODAfication  \cite{Eskander:2013:processing}, normalizing DA text into the CODA convention. We work with a unique parallel corpus of multiple Arabic dialects \cite{eryani-etal-2020-spelling}, focusing on five cities: Beirut, Cairo, Doha, Rabat, and Tunis.
    \item We benchmark newly developed pretrained sequence-to-sequence (Seq2Seq) models on the task of CODAfication. 
    \item We demonstrate that using dialect identification information improves the performance across all dialects.
\end{itemize}


Next, we discuss some related work (\S\ref{sec:related-work}) and then give a background on Arabic linguistic facts, CODA, and the data we use to train and test our models (\S\ref{sec:background}). We describe our approach in \S\ref{sec:approach} and present our experimental setup and results in \S\ref{sec:exp-setup}. 

\newpage 
\section{Related Work}
\label{sec:related-work}
\subsection{Dialectal Arabic Text Normalization}
DA NLP research has been receiving a considerable amount of attention, mainly due to the availability of monolingual and multilingual DA corpora \cite{McNeil:2011:tunisian,Zaidan:2011:arabic,Zbib:2012:machine,Cotterell:2014:multi-dialect,Salama:2014:youdacc,Jeblee:2014:domain,Al-Badrashiny:2016:lili,Zaghouani:2018:araptweet,Abdul-Mageed:2018:you,Bouamor:2019:madar}. 
While MSA has well-defined orthographic standards, none of the Arabic dialects do today.
As a result, almost all DA corpora were created without following any spelling conventions or standards, which are necessary for building robust DA NLP applications, e.g., machine translation \cite{erdmann-etal-2017-low}. To mitigate this problem, several efforts have been introduced to standardize and develop orthographic conventions for Arabic dialects. \newcite{Habash:2012:conventional} introduced the Conventional Orthography for Dialectal Arabic (CODA), the very first attempt to create guidelines and spelling conventions for Egyptian Arabic orthography. The convenience CODA offered by providing a standardized orthography led to the creation of many CODA extensions covering various dialects including Tunisian, Algerian, Palestinian, Moroccan, Yemeni, and Gulf Arabic \cite{Zribi:2014:conventional,Saadane:2015:conventional,
Jarrar:2016:curras,Turki:2016:conventional,Khalifa:2018:morphologically}. Each of these extensions tended to curate its own list of exceptional spellings for closed class words. 
\newcite{Habash:2018:unified} introduced a unified set of guidelines for Arabic Dialect
orthography -- dubbed CODA* (CODA Star). 

CODA has been used in the creation of a number of resources for DA NLP \cite{Habash:2012:morphological,Eskander:2013:processing,ARZTB:2014,Diab:2014:tharwa, Pasha:2014:madamira,Jarrar:2016:curras,Khalifa:2018:morphologically,eryani-etal-2020-spelling}.
Most relevant to this paper is the work of   
\newcite{eryani-etal-2020-spelling} who extended a portion of the  MADAR Corpus \cite{bouamor-etal-2018-madar} to create
the MADAR CODA Corpus, a collection of 10,000 sentences from five Arabic city dialects (Beirut, Cairo, Doha, Rabat, and Tunis) represented in the CODA standard in parallel with their original raw form. We use this corpus to train and test our models.  



In terms of modeling approaches to CODAfication, the first work was proposed by \newcite{Eskander:2013:processing} where they introduced CODAFY, a feature-based machine learning classifier to normalize Egyptian Arabic into CODA.
\newcite{al-badrashiny-etal-2014-automatic} and \newcite{shazal-etal-2020-unified} targeted CODA output for dialectal Arabizi (Romanized Arabic) input.
Most other approaches attempted to normalize DA texts into CODA as part of morphological analysis and disambiguation \cite{pasha-etal-2014-madamira,zalmout-etal-2018-noise,khalifa-etal-2020-morphological,zalmout-habash-2020-joint,obeid-etal-2022-camelira}. Our work is most similar to the one of  \newcite{Eskander:2013:processing} where we consider the task of CODAfication as a standalone text normalization task.

There has been some work on normalizing DA into MSA \cite{Shaalan:2007:transferring,Salloum:2011:dialectal,Salloum:2012:elissa,alnajjar:2024}. While all this work is similar to ours in that dialectal input is processed, our output is still dialectal and not in MSA. Moreover, our proposed work has some similarities to grammatical error correction (GEC) for MSA \cite{Zaghouani:2014:large,Zaghouani:2015:correction,Mohit:2014:first,Rozovskaya:2015:second,Watson:2018:utilizing,habash-palfreyman-2022-zaebuc,kwon-etal-2023-beyond,alhafni-etal-2023-advancements}. However, our task is different from GEC for MSA since GEC assumes a standard orthography that the writer is also assumed to aim for.

\subsection{Dialect Identification}

Dialect Identification (DID) is the task of identifying the dialect of a given speech or text fragment \cite{Etman:2015:language}. Since informal conversations in real-world and online settings are typically conducted in DA, there has been a growing interest in developing and scaling automatic Arabic DID systems. This can be observed in the organization of multiple shared tasks \cite{Bouamor:2019:madar,abdul-mageed-etal-2021-nadi,abdul-mageed-etal-2022-nadi,abdul-mageed-etal-2023-nadi, abdul-mageed-etal-2024-nadi} and the existence of various datasets and tools \cite{Zaidan:2011:arabic,Bouamor:2014:human,Salama:2014:youdacc,Alsarsour:2018:dart,abu-kwaik-etal-2018-shami,Zaghouani:2018:araptweet,salameh-etal-2018-fine,Bouamor:2019:madar,abdelali-etal-2021-qadi,baimukan-etal-2022-hierarchical}. Besides its obvious use for profiling \cite{rangel2019author}, DA identification has already proved to be helpful for system selection in NLP tasks such as machine translation \cite{salloum-etal-2014-sentence}, and morphological tagging \cite{obeid-etal-2022-camelira}.
In our work, we explore using text DID at the sentence-level in aiding CODAfication. For this, we use the CAMeL Tools \cite{obeid-etal-2020-camel} DID implementation of 
\newcite{salameh-etal-2018-fine}.

\begin{table*}
    \includegraphics[width=\textwidth]{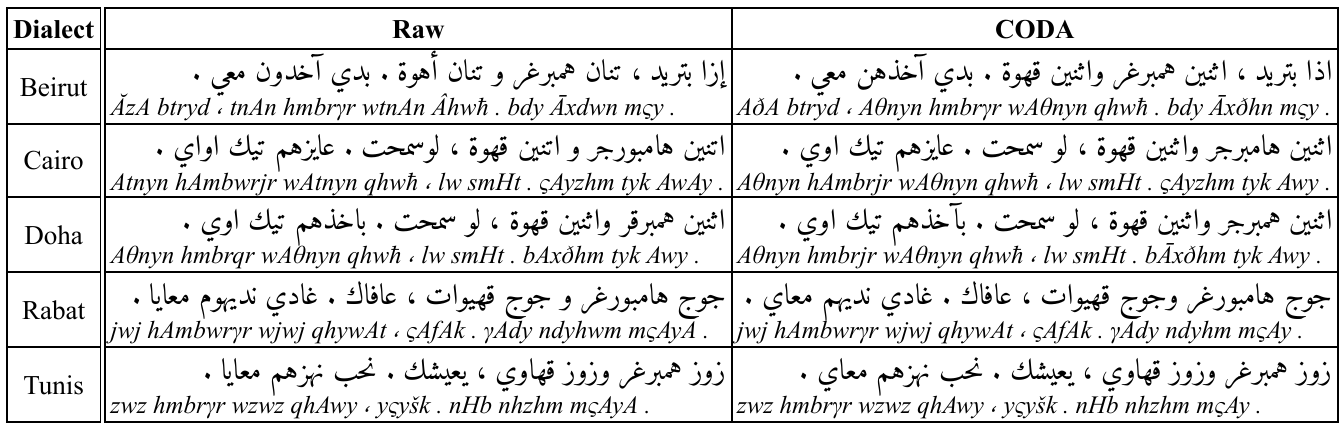}
    \caption{An example sentence from the MADAR CODA Corpus in its raw and CODA parallel forms across five city dialects. The DA sentences are provided along with their transliterations in the HSB scheme \cite{Habash2007}. The sentence in the table can be translated as \textit{``We would like two hamburgers and two coffees. To go, please.''}} 
    \label{tab:data-example} 
\end{table*}

\section{Background}
\label{sec:background}
\subsection{Arabic Linguistic Facts}

Arabic encompasses a wide range of dialectal varieties, with Modern Standard Arabic (MSA) serving as the common language of culture, media, and education across the Arab world.  However, MSA is not the native language of any Arabic speaker, as dialectal Arabic dominates daily conversations. When
native speakers write or speak (e.g., TV shows) in MSA, there is frequent
code-mixing with the dialects in terms of phonological, morphological, and lexical choices \cite{Abu-Melhim:1991:code-switching,Habash:2008:guidelines,Bassiouney:2009:arabic}. While Arabic dialects are typically classified regionally, e.g., Egyptian, North African, Levantine, and Gulf \cite{Habash:2010:introduction},  hierarchical labels have been proposed to include countries, provinces, and cities \cite{baimukan-etal-2022-hierarchical}.
In this work, we focus on five city dialects: Beirut, Cairo, Doha, Rabat, and Tunis.

Despite their similarities, DA and MSA have many differences that prevent MSA tools from being effectively utilized for dialectal text. Arabic dialects vary phonologically, lexically, and morphologically from MSA and from each other; and they vary from region to region and, to a lesser extent, from city to city in each region \cite{Watson:2007:phonology}. While MSA has a well-defined standard orthography, none of the Arabic dialects do today. When Arabic speakers write in DA, they typically write in a way that reflects the phonology or the etymology of the words. Therefore, apart from unintentional typographical errors, no spelling of a dialectal word can be deemed truly ``incorrect''. This phenomenon is referred to as \textit{spontaneous orthography} \cite{Eskander:2013:processing,eryani-etal-2020-spelling}. 
%
%
For instance, the word for `small [feminine singular]' in the Beirut dialect, \textipa{/zKi:ri/}, can be written in a range of spontaneous Arabic spellings, some of which highlighting its phonology and others its etymological connections to MSA  \<صغيرة> \textit{S{\GAYN}yr{\TAMARBUTA}} \textipa{/s\textsuperscript{\textrevglotstop}aKi:ra[t]/}. These include: 
\<زغيري>~\textit{z{\GAYN}yry},
\<زغيره>~\textit{z{\GAYN}yrh},
\<زغيرة>~\textit{z{\GAYN}yr{\TAMARBUTA}},
\<صغيري>~\textit{S{\GAYN}yry},
\<صغيره>~\textit{S{\GAYN}yrh}, and
\<صغيرة>~\textit{S{\GAYN}yr{\TAMARBUTA}}.



\begin{table*}[t!]
\centering 
    \includegraphics[width=\textwidth]{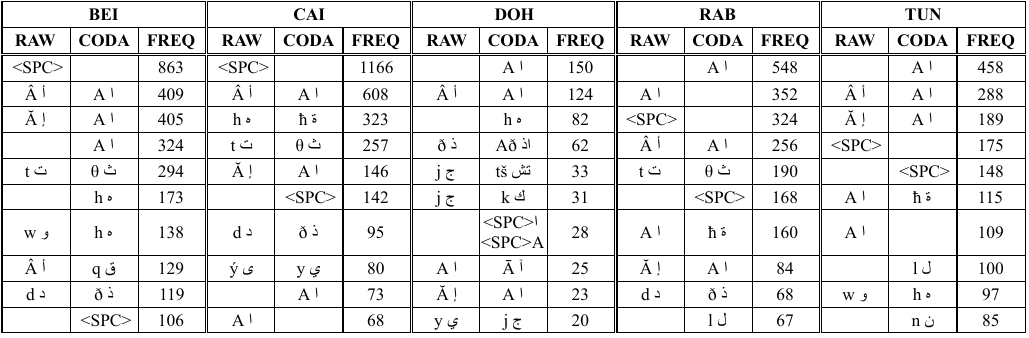}
    \caption{The top 10 character edit transformations from raw to CODA in the entire MADAR CODA dataset across the five dialects. $<$SPC$>$  indicates an explicit white space; whereas an empty cell indicates a \textit{null} string. } 
    \label{tab:coda-corpus-char-edit} 
\end{table*}

\subsection{CODA}

CODA*\footnote{Pronounced \textit{CODA Star}, as in, \textit{for any dialect}.} \cite{Habash:2018:unified}
consolidates and standardizes several prior dialect-specific CODA conventions \cite{Habash:2012:conventional,Saadane:2015:conventional,Turki:2016:conventional,Khalifa:2016:large,Jarrar:2016:curras}. 
%
CODA*, henceforth CODA, is an internally consistent and coherent convention for writing all DA varieties using the Arabic script aiming to balance dialectal uniqueness with MSA-DA similarities. 
CODA ensures consistency by controlling the natural spelling tendencies in spontaneous orthography that arise from writers considering etymological or phonological references of words.  And while it is created for computational purposes, it is designed to be easily learnable and readable.

In the example mentioned above, the Beirut dialect word 
\textipa{/zKi:ri/} `small [feminine singular]'
is written in a form reflective of MSA etymology: 
\<صغيرة>~\textit{S{\GAYN}yr{\TAMARBUTA}}. 
Other examples of CODA from the MADAR CODA Corpus \cite{eryani-etal-2020-spelling} appear in Table~\ref{tab:data-example}. Note that foreign words pose a particular challenge to CODA due to the ambiguous phonological signals in the Arabic raw text. Consequently, \newcite{eryani-etal-2020-spelling}  adopted a minimalistic strategy for CODAfying these words, resulting in some plausible but inconsistent variants. For example, the word for `hamburger' in Table~\ref{tab:data-example} appears as both \<همبرغر>~\textit{hmbr{\GAYN}r}  and 
\<همبرجر>~\textit{hmbrjr}.  




\subsection{MADAR CODA Corpus}
We use the manually annotated MADAR CODA Corpus \cite{eryani-etal-2020-spelling}, a collection of 10,000 sentences from five Arabic city dialects (Beirut, Cairo, Doha, Rabat, and Tunis) represented in the CODA standard in parallel with their original raw form. The sentences come from the Multi-Arabic Dialect Applications and Resources (MADAR) Project \cite{bouamor-etal-2018-madar} and are in parallel across the cities (2,000 sentences from each city).

The corpus is originally split into train and test, with each split consisting of 5,000 parallel sentences (1,000 per dialect). In our setup, we combine the original train and test splits and then divide the data randomly into separate training (Train), development (Dev), and testing (Test) sets. We use a 70/15/15 split, resulting in  1400, 300, and 300 sentences, respectively, per dialect. In total, we end up with 7,000 sentences for Train, 1,500 for Dev, and 1,500 for Test. Table~\ref{tab:data-example} shows an example of a sentence from the corpus in its raw and CODA parallel forms across the five city dialects. 

Table~\ref{tab:coda-corpus-char-edit} presents the top 10 character-level edit changes from raw text to CODA in the five city dialects. It is noteworthy that while there are many shared transformations, they appear with different distributions. This suggests that a model making use of DID could learn dialect-specific preferences. At the same time, the shared phenomena can aid in learning dialect-independent general patterns. 



\section{Approach}
\label{sec:approach}

We frame the CODAfication task as a controlled text generation problem. Formally, given a dialectal input sentence $X$ and its dialect $D$, the goal is to generate the CODAfied sentence $Y$ according to $P(Y | X, D)$. One way to condition text generation models on the desired dialect, $D$, is to represent it as a special ``control'' token appended to the input sequence $[D;X]$, which acts as a side constraint \cite{sennrich-etal-2016-controlling}. In Seq2Seq models, this allows the encoder to learn a representation for this token as any other token in its vocabulary, and the decoder attends to this representation to guide the generation of the output sequence. 

This simple strategy has been used in various controlled text generation tasks such as machine translation \cite{Sennrich:2016:neural,Sennrich:2016:linguistic,Johnson:2016:googles,agrawal-carpuat-2019-controlling}, style transfer \cite{niu-etal-2017-study,niu-etal-2018-multi}, text simplification \cite{yanamoto-etal-2022-controllable,agrawal-carpuat-2023-controlling}, and Arabic gender rewriting \cite{alhafni-etal-2022-user}.



\begin{table*}[t!]
\centering 
    \includegraphics[width=1.25\columnwidth]{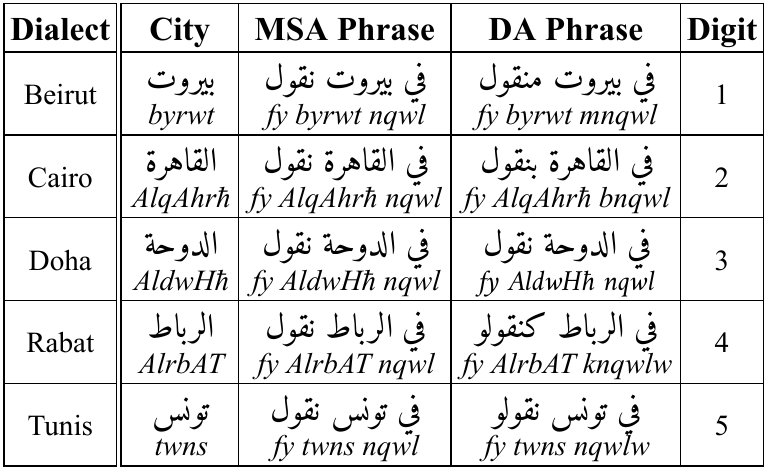}
    \caption{The four different types of control tokens we use in our experiments.} 
    \label{tab:control-tokens} 
\end{table*}

\begin{table*}[th]
    \centering
    \includegraphics[width=\textwidth]{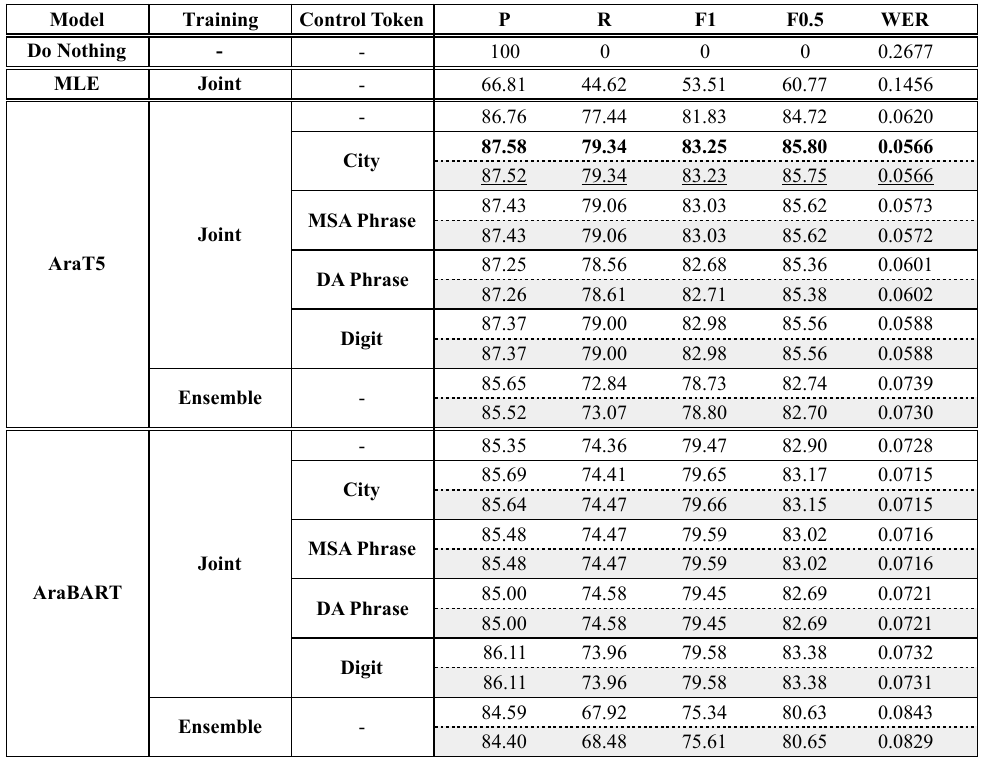}
    \caption{Results of number of systems on the Dev set. Results in grey indicate using gold DID labels (i.e., Oracle). Bolding indicates the best results. Best results in the oracle setup are underlined. } 
    \label{tab:dev-results} 
\end{table*}

We experiment with two recently developed pretrained Arabic Transformer-based Seq2Seq models: AraBART \cite{kamal-eddine-etal-2022-arabart}, which was pretrained on 24GB of MSA data primarily from the news domain, and AraT5-v2 \cite{nagoudi-etal-2022-arat5, elmadany-etal-2023-octopus}, which was pretrained on a larger dataset of 250GB covering MSA, DA, and Classical Arabic (CA) data.

We explore using four different control tokens to pass the dialect information to the models. Table~\ref{tab:control-tokens} presents the control tokens we considered in our experiments:

\begin{itemize}
    \item \textbf{City}: The name of the city where the Arabic dialect is spoken.
    \item \textbf{MSA Phrase}: An MSA phrase that follows the template \<نقول>~<\textit{city}>~\<في> `in~<\textit{city}>~we~say', where <\textit{city}> represents one of the five cities whose dialects we are modeling.

        \item \textbf{DA Phrase}: A DA phrase that follows the template <\textit{we-say}>~<\textit{city}>~\<في> `in~<\textit{city}>~we~say', where <\textit{city}> represents one of the five dialects we are modeling, and <\textit{we-say}> represents a spontaneous orthography of the dialectal version of the phrase `we say'.

    \item \textbf{Digit}: An ad hoc unique numerical value for each dialect.
\end{itemize}

During training, we use the gold dialect for each sentence to induce its control tokens. To obtain the dialect during inference, we use the DID system that is available in CAMeL Tools \cite{obeid-etal-2020-camel}. The system is an implementation of \newcite{salameh-etal-2018-fine}'s best-performing model on the MADAR shared task on DID \cite{Bouamor:2019:madar}. The system models DID for the five city dialects and MSA. We fine-tune the Seq2Seq models on a single GPU for 10 epochs, a batch size of 16, and a maximum sequence length of 200 using Hugging Face's Transformers \cite{Wolf:2019:huggingfaces}. We use learning rates of 5e-5 and 1e-4, for AraBART and AraT5, respectively. During inference, we use beam search with a beam width of 5.

\begin{table*}[th]
    \centering
    \includegraphics[width=\textwidth]{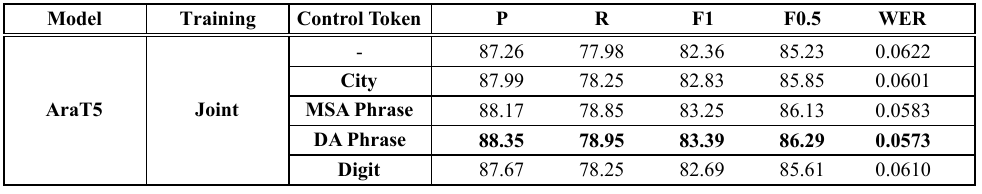}
    \caption{Results of the AraT5 variants on the Test set.} 
    \label{tab:test-results} 
\end{table*}

\begin{table*}[ht!]
    \centering
    \includegraphics[width=0.92\textwidth]{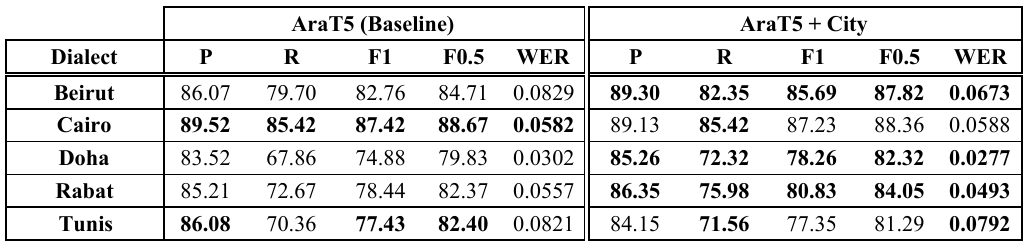}
    \caption{Dialect-specific results of the best system (AratT5 + City) against the baseline (AraT5) on the Dev set.} 
    \label{tab:dev-results-dialect} 
\end{table*}

\begin{table*}[ht!]
    \centering
    \includegraphics[width=0.92\textwidth]{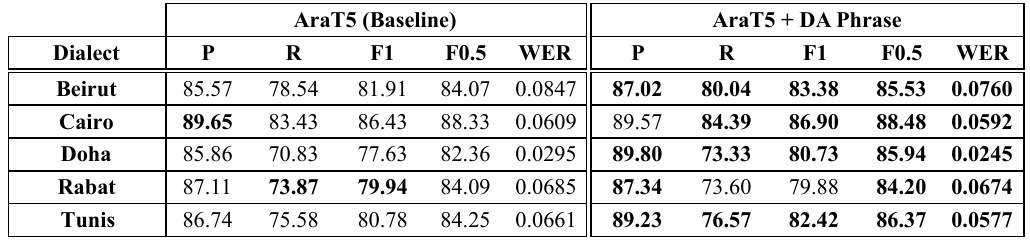}
    \caption{Dialect-specific results of the best system (AratT5 + DA Phrase) against the baseline (AraT5) on the Test set.} 
    \label{tab:test-results-dialect} 
\end{table*}

\section{Experimental Setup and Results}
\label{sec:exp-setup}
\subsection{Metrics}

We use the MaxMatch (M\textsuperscript{2}) scorer \cite{dahlmeier-ng-2012-better}, which is predominantly used to evaluate grammatical error correction systems. The M\textsuperscript{2} scorer assesses the edits made by the system against the `gold standard' edits in the target CODA, calculating precision (P), recall (R), F\textsubscript{1}, and  F\textsubscript{0.5} scores. F\textsubscript{0.5} weighs precision twice as much as recall, to prioritize the accuracy of edits relative to all edits made by the system. To obtain the gold edits, we use the alignment algorithm that was proposed by \newcite{alhafni-etal-2023-advancements}. We also use their optimized version of the M\textsuperscript{2} scorer that deals with the extreme running times of the original release in cases where the generated outputs differ significantly from the input.

Moreover, and to be consistent with previous work, we report the Word Error Rate (WER). However, we believe that WER is not a suitable metric for the task of CODAfication due to the high similarity between the input and output sentences.

\subsection{Models: Baselines and Systems}

\paragraph{Do Nothing} Our first baseline simply copies the input sentences to the output. This baseline highlights the level of similarity between the inputs and outputs.

\paragraph{Maximum Likelihood Estimation (MLE)} For the second baseline, we build a simple word-level lookup model to map input words to their CODAfied versions. We first obtain word-level alignments over all the training data from all the dialects (\textbf{Joint}) by using the algorithm developed by \newcite{alhafni-etal-2023-advancements}. We then exploit the alignments to implement the lookup model as a bigram maximum likelihood estimator: given an input word with its bigram surrounding context ($w_i$, $w_{i-1}$), and a CODAfied target word ($y_{i}$), the model is built by
computing $P(y_i | w_i , w_{i-1})$ over the training examples. During inference, we generate all possible alternatives for the given input word ($w_i$). If the bigram context ($w_i$, $w_{i-1}$) was not observed in the training data, we backoff to a unigram context.
If the input word was not observed during training, we pass it to the output as it is.


\paragraph{Seq2Seq} We train both AraBART and AraT5 on all the dialects' training data jointly with and without using DID information. We refer to this modeling setup as \textbf{Joint}. Moreover, to examine the effect of the joint dialectal training, we train five separate models, one for each dialect. During inference, we combine the separate models in an ensemble setup where we use the DID predictions for each sentence to select the appropriate model. We refer to this setup as \textbf{Ensemble}.

\subsection{Results}
\paragraph{Overall Results} Table~\ref{tab:dev-results} presents the results on the Dev set. Among the baselines, both AraBART and AraT5 demonstrate superior performance compared to the MLE model. In terms of training setups, \textbf{Joint} training outperforms \textbf{Ensemble} models for both AraBART and AraT5, with AraT5 being the better performer achieving 84.72 F\textsubscript{0.5}. 

When we train the AraBART \textbf{Joint} variants with DID control tokens, the performance increases compared to the AraBART \textbf{Joint} baseline, except when training with the \textbf{DA Phrase} DID control token. All the AraT5 \textbf{Joint} variants benefit from training with DID control tokens compared to the AraT5 baseline, with the \textbf{City} control token being the best performer with 85.80 F\textsubscript{0.5} (1.08 increase over the AraT5 baseline and statistically significant at $p < 0.05$).\footnote{Statistical significance was done using a two-sided approximate randomization test.} It is noteworthy that the AraT5 variants perform better compared to their AraBART counterparts across all experiments. We suspect this is due to the fact the data used to pretrain AraT5 consisted of a mix of MSA, DA, and CA compared to only MSA in the case of AraBART's pretraining. 

Since AraT5 performed better than AraBART across all experiments, we present the results on the Test set using AraT5 and its variants in Table \ref{tab:test-results}. Training AraT5 with the \textbf{DA Phrase} control token yields the best performance on the Test set with 86.29 F\textsubscript{0.5} (1.06 increase over the AraT5 baseline and statistically significant at $p < 0.05$). 

\paragraph{DID Efficacy} We estimate an oracle upper bound by using gold DID labels during inference on the Dev set (Table~\ref{tab:dev-results}). We do not notice significant improvements across all variants compared to the models that use predicted DID labels. In some cases, using gold DID labels results in identical performance to models using predicted labels. This can be attributed to the robustness of our CODAfication models and the reliability of the DID system we are using, which achieves a high accuracy of 92.1\% on the Dev set. 

Most of the prediction errors made by the DID system occur in sentences lacking distinctive cues that would allow clear assignment to a specific dialect. Therefore, these errors cannot be considered true errors, but rather stem from the MADAR dataset's limitation of not having multi-dialectal labels. This is consistent with the findings of \newcite{keleg-magdy-2023-arabic} where they manually analyzed the errors of a single-label DID system and found that $\sim$66\% of the errors are not true errors 
and could be resolved with multi-dialect labels.

\paragraph{Dialect-Specific Results} We present the dialect-specific results on the Dev and Test sets in Tables~\ref{tab:dev-results-dialect} and \ref{tab:test-results-dialect}, respectively. Our best system on the Dev set, AraT5 trained with the \textbf{City} DID control token, improves the results over the AraT5 baseline for all dialects (with the largest increase seen for Beirut at 3.11 F\textsubscript{0.5}), except for Cairo and Tunis, where the performance drop is attributed to decreased precision rather than recall. This suggests that our best system may be making unnecessary extra rewrites. On the Test set, our best system, AraT5 trained with the \textbf{DA Phrase} DID control token, improves the results over the AraT5 baseline across all dialects, with the largest increase for Doha at 3.58 F\textsubscript{0.5}.

\begin{table}[ht!]
\centering
\small
\setlength{\tabcolsep}{4pt}
\begin{tabular}{|l|c|c|c|}
\hline
\textbf{Category} & \textbf{\%} & \textbf{Error} & \textbf{CODA} \\ \hline\hline
\textbf{Non-CODA} & 46\% & \<تحدست> & \<تحدثت>\\
&& \textit{tHdst} & \textit{tHd{\THA}t} \\ \hline
\textbf{Hallucination} & 19\% & \<ديقة.> & \<دقيقة.>\\ 
&& \textit{dyq{\TAMARBUTA}.} & \textit{dqyq{\TAMARBUTA}.} \\ \hline
\textbf{Valid} & 13\% & \<هامبورجر> & \<هامبرجر>\\ 
&& \textit{hAmbwrjr} & \textit{hAmbrjr} \\ \hline
\textbf{Deletion} & 9\% & \<اوصلة> &
\<اوصل له> \\
&& \textit{AwSl{\TAMARBUTA}} & \textit{AwSl lh} \\ \hline
\textbf{Related} & 9\% & \<شرف> & \<الشرف>\\ 
\textbf{Hallucination} && \textit{{\SHIN}rf} & \textit{Al{\SHIN}rf} \\ \hline
\textbf{Punctuation} & 4\% & \<فاتتني>"""""""
& \<فاتتني>"""\\

&& \textit{fAttny"""""""} & \textit{fAttny"""} \\ \hline
\end{tabular}
\caption{Distribution of errors in the Dev set with one example per error type.}
\label{tab:error_distribution}
\end{table}

\subsection{Error Analysis}


To gain insights into the errors present in our best performing system on the Dev set, we conducted an error analysis on a sample of 100 cases, which accounted for 21\% of the total 471 erroneous instances in the generated output. 
We classified these errors into specific categories, with results and examples provided in Table~\ref{tab:error_distribution}:

\begin{itemize}
    \item \textbf{Non-CODA}: These are cases characterized by having plausible spontaneous spelling but incorrect CODA. This is the largest group of errors.
    \item \textbf{Hallucination and Related Hallucination}: Hallucinations refer to word rewrites that are implausible under any circumstance as a CODA correction or non-CODA spelling. We distinguish cases that seem morphologically related to the input but are actually unrelated forms.  We observe that 2/3 of the cases were largely unrelated to the reference.  
    \item \textbf{Valid}: This category encompasses valid alternative spellings, particularly those associated with proper nouns and foreign words.
    \item \textbf{Deletion}: Deletions refer to omitted words. 55.6\% of these are non-CODA spellings, e.g., a missed split (Table~\ref{tab:error_distribution} example), while the rest are divided between gold errors and hallucinations.
    \item \textbf{Punctuation}:  Punctuation generation errors.  
\end{itemize}

The error analysis highlights that CODA issues constitute a significant portion of the remaining errors, potentially accounting for half of the cases between non-CODA words and deletions. Hallucinations, whether minor or severe, make up nearly a third of the errors. These findings suggest the need for more training data and improved models to address these problems. The presence of valid variants, which represent one-eighth of the errors, indicates the need to adopt a multi-reference approach for text normalization evaluation.



\section{Conclusion and Future Work}
\label{sec:conclusion}

We explored and reported on the task of CODAfication, i.e., normalizing Dialectal Arabic into the Conventional Orthography for Dialectal Arabic (CODA). We benchmarked newly developed pretrained Seq2Seq models on the task of CODAfication. We further showed, for the first time to our knowledge,  that using dialect identification information improves Arabic text normalization.

In future work, we plan to explore other modeling approaches, including multitask learning models for both DID and CODAficiation, 
as well as text editing models \citep[\textit{inter alia}]{omelianchuk-etal-2020-gector}.
We also plan to extend our work to CODA data sets for other dialects 
\cite{Jarrar:2016:curras,Khalifa:2016:large}, evaluate the added value of improved CODAfication on downstream NLP tasks, and develop models of CODA error type classification \cite{belkebir-habash-2021-automatic}.

\section*{Limitations}
%

Although we benchmarked pretrained Seq2Seq models on the task of CODAfication and demonstrated the added improvements of using dialect identification information, we did not conduct experiments to showcase the added value of the task of CODAfication on downstream NLP tasks such as sentiment analysis and machine translation. The efficacy of CODAfication in enhancing these downstream applications, particularly with newer models, remains an area for future exploration.

Our work is based on a unique curated parallel corpus encompassing multiple Arabic dialects from five cities. While this dataset provides valuable insights into CODAfication performance across diverse dialectal variations, it also introduces limitations in generalizing our findings to a broader spectrum of Arabic dialects beyond our specific dataset. Future research should aim to extend the evaluation of CODAfication models across a more extensive range of dialectal datasets to ensure robustness and applicability across diverse linguistic contexts.

\section*{Acknowledgements}
 We acknowledge the support of the High Performance Computing Center at New York University Abu Dhabi.  
 We thank the anonymous reviewers for their feedback and suggestions.

\bibliography{custom,anthology,camel-bib-v3}




\end{document}